\documentclass[10pt, a4paper]{article}
\usepackage{lrec}
\usepackage{graphicx}
\usepackage{tabularx}
\usepackage{soul}
\usepackage{hyperref}
\usepackage{url}
\usepackage{booktabs}

\usepackage{epstopdf}
\usepackage[latin1]{inputenc}

\usepackage{xstring}

\title{Challenge Dataset of Cognates and False Friend Pairs from Indian Languages\\ \vspace*{.5\baselineskip} }

\name{Diptesh Kanojia\textsuperscript{$\dagger$,$\clubsuit$,$\star$}, Pushpak Bhattacharyya\textsuperscript{$\dagger$}, Malhar Kulkarni\textsuperscript{$\dagger$}, and Gholamreza Haffari\textsuperscript{$\star$}}

\address{\textsuperscript{$\dagger$}Indian Institute of Technology Bombay, India\\
\textsuperscript{$\clubsuit$}IITB-Monash Research Academy, India\\
\textsuperscript{$\star$}Monash University, Australia\\
        \textsuperscript{$\dagger$}\{diptesh, pb, malhar\}@iitb.ac.in, \textsuperscript{$\star$}gholamreza.haffari@monash.edu\\}

\abstract{
Cognates are present in multiple variants of the same text across different languages (\textit{e.g.,} ``hund'' in German and ``hound'' in English language mean ``dog''). They pose a challenge to various Natural Language Processing (NLP) applications such as Machine Translation, Cross-lingual Sense Disambiguation, Computational Phylogenetics, and Information Retrieval. A possible solution to address this challenge is to identify cognates across language pairs. In this paper, we describe the creation of two cognate datasets for twelve Indian languages, namely Sanskrit, Hindi, Assamese, Oriya, Kannada, Gujarati, Tamil, Telugu, Punjabi, Bengali, Marathi, and Malayalam. We digitize the cognate data from an Indian language cognate dictionary and utilize linked Indian language Wordnets to generate cognate sets. Additionally, we use the Wordnet data to create a False Friends' dataset for eleven language pairs. We also evaluate the efficacy of our dataset using previously available baseline cognate detection approaches. We also perform a manual evaluation with the help of lexicographers and release the curated gold-standard dataset with this paper. \\ \newline \Keywords{cognate sets, Indian languages, cognate dataset, true cognates, false friends, gold data} 
}

\begin{document}

\maketitleabstract

\section{Introduction and Motivation}

Cognates are words that have a common etymological origin. For \textit{e.g.,} the French and English word pair, \textit{Libert{\'e} - Liberty}, reveals itself to be a cognate through orthographic similarity. Automatic Cognate Detection (ACD) is the task of detecting such etymologically related words or word sets among different languages\footnote{Cognates can also exist in the same language. Such word pairs/sets are commonly referred to as \textit{doublets}.}. They share a formal and/or semantic affinity. Cognate words can facilitate the Second Language Acquisition (SLA) process, particularly between related languages. They have similar meanings and, therefore, can support the acquisition and/or learning of a non-native language.   However, although they can accelerate vocabulary acquisition, learners also have to be aware of false friends and partial cognates. False friends are similar words that have distinct, unrelated meanings. For example, \textit{``gift''} in German means \textit{``poison''} unlike its English meaning. We illustrate the occurrence of one such example each for cognates and false friends' for a pair of Indian languages in Table \ref{tab:examples}. In other cases, there are partial cognates \textit{i.e.,} similar words that have a common meaning only in some contexts. For example, the word \textit{``police''} in French can translate to \textit{``police''}, \textit{``policy''} or \textit{``font''}, depending on the context. Dictionaries often include information about cognates and false friends, and there are dictionaries \cite{hammer1976english,prado1993ntc} exclusively devoted to them. \\

\begin{table}[]
\centering
\resizebox{\columnwidth}{!}{%
\begin{tabular}{@{}ccccc@{}}
\toprule
 & \textbf{Hindi (Hi)} & \textbf{Marathi (Mr)} & \textbf{Hindi Meaning} & \textbf{Marathi Meaning} \\ \midrule
\textbf{Cognate} & ank & ank & Number & Number \\
\textbf{False Friend} & shikshA & shikshA & Education & Punishment \\ \bottomrule
\end{tabular}%
}
\caption{An example each of a cognate pair and a false friend pair from the closely related Indian languages Hindi (Hi) and Marathi (Mr)}
\label{tab:examples}
\end{table}

In this paper, we describe the creation of three cognate datasets. First, we describe the digitization of one such Cognate dictionary named, ``Tatsama Shabda Kosha'' and its annotation with linked Wordnet IDs. With the help of a lexicographer, we perform the digitization of this dictionary. Further, we annotate the cognate sets from the dictionary with Wordnet synset IDs based on manual validation, where the lexicographer checks each Wordnet in the existing linked sense.Based on \newcite{kanojia2019utilizing}'s approach, we use linked Indian Wordnets to generate true cognate data and create another cognate dataset. Additionally, we use the same Wordnet data to produce a list of False Friends and release\footnote{\href{https://github.com/dipteshkanojia/challengeCognateFF}{Released Data: Github Link}} all the three datasets publicly. Our cognate sets can be utilized for lookup in phrase tables produced during Machine Translation to assess the quality of the translation system in question. They can be utilized as candidate translations for words, and our false friends' list can be utilized by language learners to avoid pitfalls during the acquisition of a second language. False Friend and Cognate detection techniques can use these lists to train automatic cognate detection approaches for Indian languages. 

\begin{figure*}[ht]
    \centering
    \includegraphics[width=\textwidth]{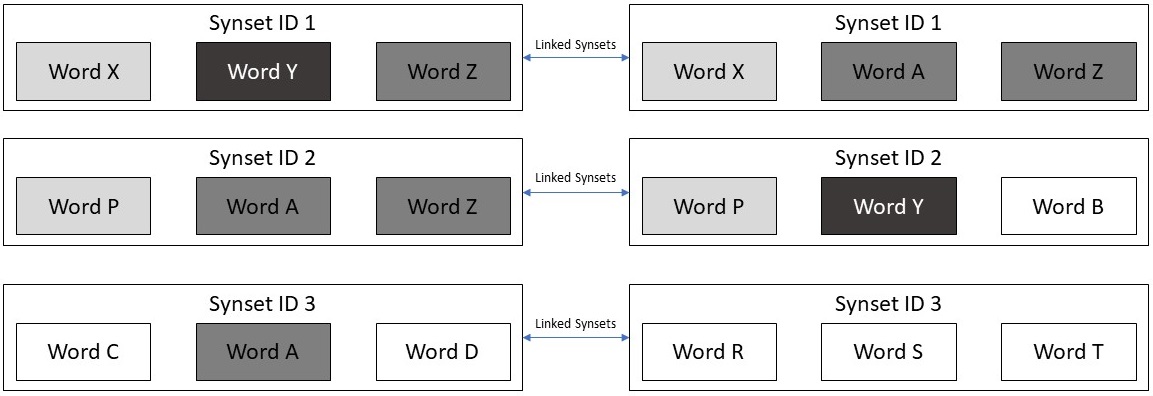}
    \caption{The difference between True Cognates (Word X and Word P), False Friends (Word Y) and Partial Cognates (Word A and Word Z) explained for creating our Datasets (D2 and D3).}
    \label{fig:cognates}
    \vspace{-2mm}
\end{figure*}

Automatic Cognate Detection (ACD) techniques help phylogenetic inference by helping isolate diachronic sound changes and thus detecting the words of a common origin \cite{rama2014gap}. The Indo-Aryan and Dravidian language families prevalent in South Asia are examples of language families with a few ancestors (Sanskrit/Persian and Proto-Dravidian, respectively). Indian language pairs borrow a large number of cognates and false friends due to this shared ancestry. Knowing and utilising these cognates/false friends can help improve the performance of computational phylogenetics \cite{rama2018automatic} as well as cross-lingual information retrieval \cite{meng2001generating} in the Indian setting, thus encouraging us to investigate this problem for this linguistic area\footnote{The term linguistic area or Sprachbund \cite{emeneau1956india} refers to a group of languages that have become similar in some way as a result of proximity and language contact, even if they belong to different families. The best-known example is the Indian (or South Asian) linguistic area.}. Some other applications of cognate detection in NLP have been sentence alignment \cite{simard1993using,melamed1999bitext}, inducing translation lexicons \cite{mann2001multipath,tufis2002cheap}, improving statistical machine translation models \cite{al1999statistical}, and identification of confusable drug names \cite{kondrak2004identification}. All these applications depend on an effective method of identifying cognates by computing a numerical score that reflects the likelihood that the two words are cognates. Our work provides cognate sets for Indian languages, which can help the automated cognate detection methodologies and can also be used as possible translation candidates for applications such as MT. 

\section{Related Work}

\newcite{wu2018creating} release cognate sets for Romance language family and provide a methodology to complete the cognate chain for related languages. Our work releases similar data for Indian languages. Such a cognate set data has not been released previously for Indian languages, to the best of our knowledge. Additionally, we release lists of false friends' for language pairs. These cognates can be used to challenge the previously established cognate detection approaches further. \newcite{kanojia2019cognate} perform cognate detection for some Indian languages, but a prominent part of their work includes \textit{manual verification and segratation} of their output into cognates and non-cognates. Identification of cognates for improving IR has already been explored for Indian languages \cite{makin2007approximate}. Orthographic/String similarity-based methods are often used as baseline methods for cognate detection, and the most commonly used method amongst them is the Edit distance-based similarity measure \cite{melamed1999bitext}. 

Research in automatic cognate detection using various aspects involves computation of similarity by decomposing phonetically transcribed words \cite{kondrak2000new}, acoustic models \cite{mielke2012assessing}, clustering based on semantic equivalence \cite{hauer2011clustering}, and aligned segments of transcribed phonemes \cite{list2012lexstat}. \newcite{rama2016siamese} employs a Siamese convolutional neural network to learn the phonetic features jointly with language relatedness for cognate identification, which was achieved through phoneme encodings. \newcite{jager2017using} use SVM for phonetic alignment and perform cognate detection for various language families. Various works on orthographic cognate detection usually take alignment of substrings within classifiers like SVM \cite{ciobanu2014automatic,ciobanu2015automatic} or HMM \cite{bhargava2009multiple}. \newcite{ciobanu2014automatic} employ dynamic programming based methods for sequence alignment. Among cognate sets, common overlap set measures like set intersection, Jaccard \cite{jarvelin2007s} or XDice \cite{brew1996word} could be used to measure similarities and validate the members of the set. 

\begin{table*}[ht]
\centering
\resizebox{\textwidth}{!}{%
\begin{tabular}{|c|c|c|c|c|c|c|c|c|c|c|c|}
\hline
\textbf{Language Pair}     & \textbf{Hi-Bn} & \textbf{Hi-Gu} & \textbf{Hi-Mr} & \textbf{Hi-Pa} & \textbf{Hi-Sa} & \textbf{Hi-Ml} & \textbf{Hi-Ta} & \textbf{Hi-Te} & \textbf{Hi-As} & \textbf{Hi-Kn} & \textbf{Hi-Or} \\ \hline
\textbf{Potential Candidates}             & $50959$        & $81834$        & $47718$        & $25044$        & $33921$        & $18084$        & $5203$         & $16230$        & $14240$        & $12480$        & $54014$        \\ \hline
\textbf{Cognates (D2)}          & 15312          & 17021          & 15726          & 14097          & 21710          & 9235           & 3363           & 936            & 3478           & 4103          & 11894          \\ \hline
\textbf{Percent Agreement} & 0.9877         & 0.9849         & 0.9838         & 0.9754         & 0.9617         & 0.9223         & 0.9033         & 0.9553         & 0.9167         & 0.9122         & 0.8833         \\ \hline
\textbf{Cohen's kappa}     & 0.7851         & 0.7972         & 0.8628         & 0.7622         & 0.7351         & 0.7046         & 0.6436         & 0.7952         & 0.7591         & 0.7953         & 0.8333         \\ \hline
\end{tabular}
}
\caption{Number of Potential Cognates, Number of cognates retained on both annotators' agreement [Cognates (D2)], Percent agreement among the annotators and Cohen's kappa score for each language pair in our dataset}
\label{tab:d2stats}
\end{table*}

\begin{table*}[]
\centering
\resizebox{\textwidth}{!}{%
\begin{tabular}{|c|c|c|c|c|c|c|c|c|c|c|c|}
\hline
\textbf{Language Pair} & \textbf{Hi-Bn} & \textbf{Hi-Gu} & \textbf{Hi-Mr} & \textbf{Hi-Pa} & \textbf{Hi-Sa} & \textbf{Hi-Ml} & \textbf{Hi-Ta} & \textbf{Hi-Te} & \textbf{Hi-As} & \textbf{Hi-Kn} & \textbf{Hi-Or} \\ \hline
\textbf{Potential Candidates} & 11128 & 10378 & 14430 & 9062 & 9285 & 5192 & 1018 & 7149 & 9374 & 3384 & 5011 \\ \hline
\textbf{False Friends (D3)} & 4380 & 6204 & 5826 & 4489 & 2193 & 1076 & 783 & 699 & 3872 & 926 & 2602 \\ \hline
\textbf{Percent Agreement} & 0.8912 & 0.9122 & 0.9233 & 0.9500 & 0.9018 & 0.8125 & 0.9288 & 0.8492 & 0.8825 & 0.9367 & 0.9133 \\ \hline
\textbf{Cohen's kappa} & 0.8827 & 0.8245 & 0.7815 & 0.9255 & 0.9452 & 0.9064 & 0.7244 & 0.8901 & 0.8432 & 0.8167 & 0.9548 \\ \hline
\end{tabular}%
}
\caption{Number of Potential False Friends, Number of False Friend pairs retained on both annotators' agreement [False Friends (D3)], Percent agreement among the annotators and Cohen's kappa score for each language pair in our dataset}
\label{tab:fftable}
\end{table*}

\begin{table}[]
\centering
\resizebox{\columnwidth}{!}{%
\begin{tabular}{@{}ccccc@{}}
\toprule
 & \textbf{Nouns} & \textbf{Verbs} & \textbf{Adjectives} & \textbf{Adverbs} \\ \midrule
\textbf{D1} & 78.20 & 0.06 & 19.00 & 0.60 \\
\textbf{D2} & 76.35 & 2.41 & 20.11 & 1.10 \\ \bottomrule
\end{tabular}%
}
\caption{The percentage share of parts-of-speech categories in cognate datasets D1 and D2}
\label{tab:pospercent}
\end{table}

\section{Dataset Creation}

We create three different datasets to help the NLP tasks of cognate and false friends' detection. In this section, we describe the creation of these three datasets for twelve Indian languages, namely Sanskrit, Hindi, Assamese, Oriya, Kannada, Gujarati, Tamil, Telugu, Punjabi, Bengali, Marathi, and Malayalam.

\subsection{D1 - True Cognate Sets}

The first dataset is created with the help of manual annotation. We digitize the book ``Tatsama Shabda Kosh'' with the help of a lexicographer. The dictionary is a collection of cognates from 15 Indian languages, but focus our work on 12 languages due to the unavailability of Wordnets for the rest of the languages. The lexicographer then also annotates each cognate set with a Wordnet sense ID after manual validation of each cognate in the twelve linked Wordnets. This helps us capture an appropriate sense for the cognate word provided via the dictionary. The annotation was performed manually with the data collected in a CSV format in a text editor. By definition, cognates are supposed to spell and mean the same. Our manual annotation using the Wordnet IDs helps provide an appropriate sense to each cognate set in the dataset and thus can help automatic cognate detection techniques utilize the synset information.

This dataset consists of 1021 cognate sets with a total of 12252 words. The book consisted of a total of 1556 cognate sets, but during manual validation, 535 were found to be partial cognates and have been ignored from this dataset. The percentage share of parts-of-speech categories for Wordnet annotated cognates released is shown in Table \ref{tab:pospercent}. Partial cognates, as previously explained, mean differently in different contexts. NLP tasks such as Machine Translation will benefit the most from gold-standard translation candidates. Keeping the application of our dataset in mind, we ignore the inclusion of partial cognates from this dataset.

\subsection{D2 - True Cognate Pairs via IndoWornet}

In their paper, \newcite{kanojia2019utilizing} identify IndoWordnet \cite{bhattacharyya2017indowordnet} as a potential resource for the task of cognate detection. They utilize deep neural network based approaches to validate their approach for cognate detection. We build this dataset using a simple orthographic similarity based approach from the IndoWordnet dataset. Our approach combines Normalized Edit Distance (NED) \cite{nerbonne1997measuring} and Cosine Similarity (CoS) \cite{salton1988term} between words. We compare synset words from every language pair using NED and populate a list of cognate sets where NED score is 0.7 and above. Similarly, we populate another list of cognate sets from every language pair using a shingle (n-gram) based Cosine Similarity with the same threshold. Due to the different methods using which NED and CoS similarity techniques compute scores, both NED and CoS output a different number of word pairs. We choose a common intersection of cognate pairs from among both the lists, and populate a final `potential cognate set' for eleven Indian language pairs. We take the help of two lexicographers and manually validate this output. We are aided by two lexicographers for each of the language pairs of Hindi (source) - (target) `other Indian languages'\footnote{We intended to isolate the lexicographers of clues from other language cognate pairs. Hence, we create cognate data in language pairs.} Each lexicographer was requested to annotate whether the given word pair is cognate or not, given the Wordnet synset information, which contained the definition of the concept and an example sentence. We retain in the final dataset, only cognate pairs, which were marked to be true cognates by both annotators. We provide the language pair wise cognate data statistics, percent agreement, and Cohen's Kappa (IAA) values for the lexicographers' annotation in Table \ref{tab:d2stats}. The percentage share of parts-of-speech categories for Wordnet annotated cognates released is shown in Table \ref{tab:pospercent}.

\subsection{D3 - False Friends' Pairs}

The creation of such a False Friends' dataset is another one of our novel contributions in this paper. We search for false friend candidate pairs by searching for commonly spelled words through the non-parallel synsets. These candidate pairs initially included partial cognates as well, since words which are commonly spelled and belong to different senses, could occur in both the contexts. We further prune this list by ensuring that these commonly spelled words do not occur in parallel synsets and also do not occur in the corresponding linked synset on either the source or the target side. Figure \ref{fig:cognates} explains our heuristic where Word Y is a False Friend among Synsets 1 and 2, Word X and Word P are True Cognates chosen for D2, and Word A / Word Z are ignored because they are partial cognates. Once we find out such unique False Friend pairs, which are exact matches in spelling but do not occur in parallel synsets, on either side, we populate our list of false friend pairs. We populate this list for eleven language pairs where Hindi is always the source language. Please note that false friends do not follow transitivity, i.e., if A and B are false friend pairs in languages X and Y, and A and C are false friends in X and Z, it is not necessary that B and C would be false friends. Hence, we populate eleven different false friend lists and take lexicographers help for each language pair to manually validate this output. Post-manual validation we choose retain the false friend pair which were annotated as false friends' by both the annotators. We report the statistics for D3 in Table \ref{tab:fftable}, which include the number of potential false friend candidates, False friend pairs after manual validation of these potential candidates, Percent agreement among both the annotators and Cohen's Kappa (IAA) score for the annotation performed.

\begin{table*}[ht!]
\centering
\resizebox{\textwidth}{!}{%
\begin{tabular}{|c|c|c|c|c|c|c|c|c|c|c|c|}
\hline
\textbf{Approaches} & \textbf{Hi-Bn} & \textbf{Hi-As} & \textbf{Hi-Or} & \textbf{Hi-Gu} & \textbf{Hi-Mr} & \textbf{Hi-Pa} & \textbf{Hi-Sa} & \textbf{Hi-Ml} & \textbf{Hi-Ta} & \textbf{Hi-Te} & \textbf{Hi-Kn} \\ \hline
\textbf{Orthographic Similarity} & 0.36 & 0.34 & 0.38 & 0.25 & 0.29 & 0.21 & 0.24 & 0.28 & 0.20 & 0.16 & 0.19 \\ \hline
\textbf{Phonetic Similarity} & 0.42 & 0.38 & 0.39 & 0.29 & 0.32 & 0.24 & 0.25 & 0.31 & 0.24 & 0.22 & 0.25 \\ \hline
\textbf{Rama et. al. (2016)} & 0.65 & \textbf{0.71} & 0.61 & 0.67 & \textbf{0.72} & 0.47 & 0.53 & 0.62 & \textbf{0.53} & \textbf{0.65} & 0.57 \\ \hline
\textbf{Kanojia et. al. (2019)} & \textbf{0.68} & \textbf{0.71} & \textbf{0.62} & \textbf{0.75} & \textbf{0.72} & \textbf{0.73} & \textbf{0.72} & \textbf{0.66} & \textbf{0.53} & 0.63 & \textbf{0.58} \\ \hline
\end{tabular}%
}
\caption{Results of the Cognate Detection Task (in terms of F-Scores) for D1+D2. We use the same architecture, features and hyperparameters as discussed in the papers for Rama et. al. (2016) and Kanojia et. al. (2019) and observe that these systems do not perform as well on our dataset, as claimed by the authors.}
\label{tab:resultscognate}
\end{table*}

\begin{table*}[]
\centering
\resizebox{\textwidth}{!}{%
\begin{tabular}{|c|c|c|c|c|c|c|c|c|c|c|c|}
\hline
\textbf{Language Pairs} & \textbf{Hi-Bn} & \textbf{Hi-As} & \textbf{Hi-Or} & \textbf{Hi-Gu} & \textbf{Hi-Mr} & \textbf{Hi-Pa} & \textbf{Hi-Sa} & \textbf{Hi-Ml} & \textbf{Hi-Ta} & \textbf{Hi-Te} & \textbf{Hi-Kn} \\ \hline
\textbf{Orthographic Similarity} & 0.36 & 0.45 & 0.49 & 0.51 & 0.53 & 0.44 & 0.52 & 0.24 & 0.29 & 0.30 & 0.50 \\ \hline
\textbf{Phonetic Similarity} & 0.60 & \textbf{0.66} & \textbf{0.67} & 0.62 & 0.59 & 0.69 & 0.61 & 0.54 & 0.48 & 0.50 & 0.57 \\ \hline
\textbf{Castro et. al. (2018)} & \textbf{0.66} & 0.64 & 0.59 & \textbf{0.65} & \textbf{0.69} & \textbf{0.73} & \textbf{0.72} & \textbf{0.65} & \textbf{0.52} & \textbf{0.69} & \textbf{0.64} \\ \hline
\end{tabular}%
}
\caption{Results of the False Friends' Detection Task (in terms of F-Scores) for D3. We use the same architecture, features and hyperparameters as discussed in the paper by Castro et. al. (2018) and observe that these systems do not perform as well on our False Friends' dataset.}
\label{tab:ffresults}
\end{table*}

\section{Experiment Setup for Evaluation}

We evaluate the challenge posed by our datasets using the tasks of automatic cognate detection and false friend detection, based on previously available approaches. In this section, we describe the task setup and approaches which show the challenges posed by these tasks. We also discuss how our dataset is a challenging dataset for these tasks, and better approaches are needed to tackle the problems posed by a morphologically richer dataset of cognates.\\

We combine D1 and D2 based on Wordnet Sense IDs and remove duplicates to form a single dataset of true cognates, which we evaluate through the task of cognate detection. We use various approaches to perform the cognate detection task \textit{viz.} baseline cognate detection approaches like orthographic similarity based, phonetic similarity based, phonetic vectors with Siamese-CNN based proposed by \newcite{rama2016siamese}, and deep neural network based approaches proposed by \newcite{kanojia2019utilizing}. We use the same hyperparameters and architectures, as discussed in these papers. For the Orthographic similarity based approach, we use the orthographic similarity between words as a feature. For the Phonetic similarity based approach, we compute the phonetic similarity between two words using phonetic vectors available via the IndicNLP Library\footnote{\url{https://anoopkunchukuttan.github.io/indic_nlp_library/}}. To classify cognate pairs, we use a simple feed forward neural network with the respective feature scores passed to a fully connected layer with ReLU activations, followed by a softmax layer (in the first two approaches). We replicate the best reported systems from \newcite{rama2016siamese} \textit{i.e.,} Siamese Convolutional Neural Network with phonetic vectors as features. To replicate \newcite{kanojia2019utilizing}'s approach, we use the Recurrent Neural Network architecture with a combination of Normalized Edit Distance, Cosine Similarity, and Jaro-Winkler Distance as reported in their paper. We have already discussed the manual validation of our datasets in the previous section, which allows us to create a more curated dataset. We use the computational approaches on this curated dataset post manual validation. For the training dataset, we use the data provided by \newcite{kanojia2019utilizing} and create training and validation sets with an 80-20 split. We then test the aforementioned approaches on our dataset.\\

For the False Friends detection task, since no such dataset is available for Indian languages, we annotate the data created by us with positive labels and divide it into train and test sets. We then add true cognates to the training dataset with negative labels since intuitively, they are the best candidates for misclassification due to common spellings just like false friends do, but in case of true cognates, they also mean the same. We test our false friends' dataset using similar approaches with baseline features like orthographic similarity, and phonetic similarity. We use a simple Feed Forward neural network as the classifier with the respective feature scores passed to a fully connected layer with ReLU activations, followed by a softmax layer. To ensure the learning algorithm has `context' available to decipher the meaning among false friends, we use the notion of distributional semantics and employ a word vectors based approach proposed by \newcite{castro2018high}. We use their approach to test the efficacy of our dataset and show better approaches need to be devised for morphologically richer languages.\\

Since the approach proposed by \newcite{castro2018high} requires monolingual word embeddings to be used, we train the monolingual word embeddings using the standard Wikimedia dumps\footnote{as on 15th October, 2019}. We extract text from the Wiki dumps and tokenize the data. We, then, train twelve monolingual word embedding models for each Indian language we are dealing with. In the next section, we discuss the results of our dataset evaluations.

\section{Results of Our Evaluation}

In table \ref{tab:resultscognate}, we show the results for the cognate detection task. We observe that on our combined cognate dataset, the current approaches do not perform well. These approaches have reported better performance for their own datasets. In most of the cases (Language pairs), \newcite{kanojia2019utilizing}'s approach performs better, but for the Hindi-Telugu language pair, \newcite{rama2016siamese}'s approach performs better. Although both the approaches perform the same for Hi-As, Hi-Mr, and Hi-Ta language pairs, the scores are still lower than what has been previously reported. We believe that these approaches perform well when on a limited dataset, moreover, when the dataset consists of words which are stripped on morphological inflections. NLP tasks such as Machine Translation and Cross-lingual Information Retrieval do not use synthetic data, which is stripped of morphological information. If the cognate detection task has to be a part of a pipeline aiding the NLP tasks, then approaches that perform the task should be robust enough to tackle a dataset such as ours. Hence, we claim our dataset to be a more challenging dataset, which should help develop better approaches. 

In table \ref{tab:ffresults}, we report the results for the task of False Friends' detection. We observe that the approach proposed by \newcite{castro2018high} does not perform as well as it does for Spanish and Portuguese, as reported previously. We believe that this approach inherently lacks the linguistic intuition which is needed for the false friends' detection task. Please recall that False friends are word pairs that spell the same but do not mean the same. But for the approach to perform well, monolingual embeddings may not be an appropriate feature. Cross-lingual word embeddings project monolingual word embeddings into a common space and thus should be able to decipher the `meaning' or the `sense' of two different words better, when they belong to different languages. Given the recent advancements in word representation models, cross-lingual word embedding based models should be employed for such a task. Please also note that we do not propose a new approach for the task of False friends' detection and hence do not perform any experimentation with cross-lingual word embeddings. However, \newcite{merlo2019cross} show that cross-lingual word embeddings obtained using the VecMap \cite{artetxe-etal-2016-learning} approach have shown promise and can be used to obtain a semantic comparison between two words from different languages. 

\section{Conclusion and Future Work}

In this paper, we describe the creation of a challenging dataset of true cognates which encompasses of cognates from two different sources. First, we digitize a cognate dictionary and annotate it with Wordnet Sense IDs for twelve Indian languages to create Dataset 1 (D1). We also use linked Indian Wordnets to create a true cognate dataset, as described in the paper. For both the datasets, we ensure a quality check with the help of manual validation. We report the percent agreement and Inter-annotator agreement for D1 and D2 in this paper, and for D2, we retain the cognate pairs, which were marked to be cognates by both the annotators; we were aided by two annotators for each language pair. Additionally, we release a curated list of False friends for eleven language pairs where the Hindi language is always the source, and other Indian languages are the target languages. We evaluate the efficacy of all these datasets using previously available approaches for the tasks of Cognate and False Friends' detection. We show that these approaches do not perform as well on our dataset, given the same hyperparameters and settings as described in their papers. We discuss these results in the previous section. We also believe that this work provides a challenging gold-standard dataset for the tasks for Cognate and False Friends' detection, which can also be used to aid the NLP tasks of Machine Translation, Cross-lingual Information Retrieval, and Computational Phylogenetics. We hope better approaches are developed for these tasks which can perform well on our challenge dataset.\\

In the near future, we shall include partial cognates in our dataset creation approach and release another dataset on the same repository. Partial cognates mean different given different contexts and can confuse an NLP task. Hence, we believe it is also important to have a challenging dataset for partial cognates as well which can be evaluated via the blingual bootstrapping approach described by \newcite{frunza2006automatic}. We would also like to evaluate our dataset on other NLP tasks and report its efficacy in aiding the tasks of MT, CLIR, Cross-lingual Question Answering \textit{etc.}

\section{Bibliographical References}
\label{main:ref}

\bibliographystyle{lrec}
\bibliography{lrec2020W-xample}


\end{document}